\documentclass{article}

\usepackage[T1]{fontenc}
\usepackage[utf8]{inputenc}

\usepackage{url}

\usepackage{graphicx}
\usepackage[noend,ruled,vlined,linesnumbered]{algorithm2e}

\SetCommentSty{mycommfont}

\usepackage{amsmath,amssymb,amsthm,amsfonts}

\RequirePackage{booktabs,multirow} 
\usepackage{arydshln}
\usepackage{hyperref}

\usepackage{bm}

\usepackage{xcolor}
\usepackage{xspace}
\newcommand{\ie}{\emph{i.e.}\xspace}

\DeclareMathOperator*{\argmin}{argmin}

\newcommand\HierASTiSeq{{\sc HierAsTiSeq}\xspace}

\newtheorem{example}{Example}

\usepackage{comment}

\definecolor{corgreen}{rgb}{.349,.8,.1}
\definecolor{corred}{rgb}{.8,0.1,0}

\usepackage[T1]{fontenc}
\usepackage[utf8]{inputenc}
\usepackage{fullpage}

\usepackage{url}

\usepackage{graphicx}
\usepackage[noend,ruled,vlined,linesnumbered]{algorithm2e}

\newcommand\citep[1]{\cite{#1}}
\newcommand\citet[1]{\cite{#1}}

\title{Clustering of timed sequences -- Application to the analysis of care pathways\footnote{This article has been published in the Journal of Data \& Knowledge Engineering. To cite this article, please use the following reference:  \textit{Clustering of timed sequences – Application to the analysis of care pathways, Thomas Guyet, Pierre Pinson, Enoal Gesny, Journal of Data \& Knowledge Engineering, 156:102401 (2025) doi:10.1016/j.datak.2024.102401}.}}
\author{Thomas Guyet$^\dagger$, Pierre Pinson$^\ddagger$ and Enoal Gesny$^{\dagger,\ddagger}$}
\date{
$\dagger$. AIstroSight, Inria, Hospices Civils de Lyon, Université Claude Bernard Lyon 1, 
56 Bld Niels Bohr, 69603 Villeurbanne, France\\
\href{mailto:thomas.guyet@inria.fr}{thomas.guyet@inria.fr}\\
$\ddagger$. Clinical Research Unit, AP-HP, Cochin Academic Hospital, France.
}

\begin{document}

\maketitle

\begin{abstract}
Improving the future of healthcare starts by better understanding the current actual practices in hospital settings. 
This motivates the objective of discovering typical care pathways from patient data. Revealing typical care pathways can be achieved through clustering. The difficulty in clustering care pathways, represented by sequences of timestamped events, lies in defining a semantically appropriate metric and clustering algorithms.

In this article, we adapt two methods developed for time series to {the clustering of} time{d} sequences: the drop-DTW metric and the DBA approach for the construction of averaged time sequences. These methods are then applied in clustering algorithms to propose original and sound clustering algorithms for timed sequences.
This approach is experimented with and evaluated on synthetic and real-world data.
\end{abstract}

\section{Introduction}

Timed sequences, composed of dated events, are types of data frequently encountered in various application domains such as computer log analysis, life course analysis in social science~\citep{gabadinho_analyzing_2011}, tourism trajectories~\citep{bernard2018towards} or healthcare pathway analysis~\citep{rama_aliclu_2019}.
A specificity of this type of data is that they have no natural vector representation.
In particular, they are formed by two very different dimensions: symbolic and temporal; and, on the other hand, their size generally varies from one example to another (e.g., in a healthcare pathway, some patients have more care events than others).
This difficulty in modeling data in a vector space makes it challenging to adapt standard data analysis methods (classification, clustering, or forecasting).

The adaptation of data analysis methods for timed sequences raises the question of formalizing a topology of the objects to be analyzed. 
Most data analysis methods make the assumption that objects lie in a metric space.
This topology assumption provides intuitive ways to generalize sparse and noisy observations.
Formalizing such a space for timed sequences involves defining a metric between timed sequences.
This problem is very similar to the one encountered in time series analysis (temporal data where the values are numerical, not event-based and generally with a regular and dense temporal sampling). For time series analysis, a large collection of metrics has been proposed and allows them to be analyzed~\citep{holder2024review,tavenard2020tslearn}. 
The challenges reveals to identify a metric that has the appropriate properties for the analysis task one wishes to carry out (e.g., the properties of a distance) and whose semantics will give the expected results. Intuitively, a similarity metric (resp. dissimilarity metric) should produce a high similarity value (resp. low dissimilarity value) when two examples are considered similar by domain experts.

For sequential data, few studies have focused on adapting data analysis methods. 
\cite{gay2015tess} and  \cite{Sebia2024} are based on tensor representation of sequential data, while~\cite{pmlr-v138-yu20a} or~\cite{nguyen2018sqn2vec} use embedding representations. Nonetheless, only a few methods handle event logs with timestamped events dates~\citep{rama_aliclu_2019,gabadinho_analyzing_2011}.

The TraMineR~\citep{gabadinho_analyzing_2011} software has been frequently used for epidemiological studies, for instance by~\cite{le2019categorical}. 
It enables the user to cluster or classify event-based temporal data with different metrics. 
Nonetheless, this method represents an event-based data as a state-sequence, i.e., a longitudinal sequence which associate at each instant a unique state. One advantage of this tool is to propose an end-to-end analysis of the data. 
Another end-to-end method for timed sequences is AliClu~\citep{rama_aliclu_2019} which combines the definition of a time warping metric with a hierarchical clustering algorithm. 
Contrary to TraMineR, it does not allow a change of metric. 

In addition to these end-to-end methods, some work was dedicated to metrics between temporal sequences~\citep{garreau2014metric,schneider2022probabilistic,compagnon:hal-03264242} or averaging~\citep{chen2017sequence,BesnardGuyet,tatti2012long}. Despite these proposals, metrics whose semantics can better take into account the temporal dimension are still lacking. In particular, the delays between the occurrence of events are not handled. Most current solutions generally only take into account the sequentiality in the data.

In this work, we extend the drop-DTW metric~\citep{dvornik2021drop} which was proposed to evaluate distances between sequences  taking into account both sequential aspects, by adapting Dynamic Time Warping~\citep{sakoe1978dynamic}, and by adding support for meaningless events that can be dropped.
This metric has the semantics that seem appropriate for our application case -- i.e., the clustering of care pathways, but as this metric is not a distance, it does not allow for the construction of an average sequence nor to be used in classical clustering methods such as the K-Means algorithm or hierarchical clustering.

\vspace{10pt}

The main contribution of this article is the proposal of an algorithm to compute a mean timed sequence based on drop-DTW. This algorithm is inspired by the DBA method~\citep{petitjean_global_2011} for time series. The construction of an average sequence meets both the needs of clustering algorithms (hierarchical clustering or K-means) and the needs of interpretation of sequence clusters (definition of a representative).

This method is applied on a Electronic Health Records (EHR) dataset in order to discover typical care-pathways. 
In the context of the OPTISOINS project\footnote{\url{https://www.bernoulli-lab.fr/project/optisoins/}}, we were particularly interested in patients who had undergone pulmonary resection surgery for oncological reasons. Care trajectories were reconstructed from data collected in the EHR data warehouse of the Greater Paris Hospital (AP-HP). The care pathway is a timed sequence representing the key stages of managing patient surgery from diagnosis to post-surgery treatments.

\vspace{5pt}

The reminder of this article is organized as follows. We first introduce notations related to timed sequences. Then, Section~\ref{sec:metric} introduces the adaptation of drop-DTW to timed sequences. Section~\ref{sec:average} presents the adaptation of the DBA algorithm for averaging timed sequences based on the drop-DTW. This averaging method is required in classical clustering algorithms that are briefly recalled in Section \ref{sec:clustering}.
Finally, Sections~\ref{sec:results} and \ref{sec:usecase}
present experiments and results on synthetic data and our use case (OPTISOINS project) respectively.

\section{Times Sequences and Probabilistic Timed Sequences}\label{sec:timed_sequences}
In this section, we introduce notations for timed sequences and probabilistic timed sequences. The probabilistic version of timed sequences enables to adapt the metrics defined for time series to timed sequences (see next section).

Let $\Sigma$ be a finite set of $n$ event types.
A timed sequence is a sequence of pairs $S=\langle (s_i,t_i) \rangle$ where $s_i\in\Sigma$ is an event type and $t_i\in\mathbb{R}$ is the event's timestamp. $S[i]=(s_i,t_i)$ denotes the $i$-th event in the sequence. $|S|$ is the length of the timed sequences, i.e. its number of events.
By convention, events in a sequence are ordered by their timestamps. Several events may occur at the same time only if their event types are different, i.e. $(s_i,t)=(s_j,t) \Rightarrow s_i=s_j$.

We propose embedding the space of timed sequences into a larger space where events are represented by their one-hot encoding, which is a vector representation. Instead of representing a sequence of events, we represent a sequence of event types distributions:
$$\langle ( [d^1_1, \dots d^n_1], t_1), \dots, ( [d^1_K, \dots d^n_K], t_K) \rangle$$
where $\bm{d}\cdot=[d^1_\cdot, \dots, d^n_\cdot]\in [0,1]^n$, such that $\sum_i d^i_\cdot=1$.
$\bm{d}_\cdot$ can be seen as a probability distribution of event types.
Thus, events can be represented in a vector space and the representation of a timed sequence becomes close to the representation of a (multivariate, irregular) time series.

\begin{figure}[tb]
\centering
\includegraphics[width=.7\textwidth]{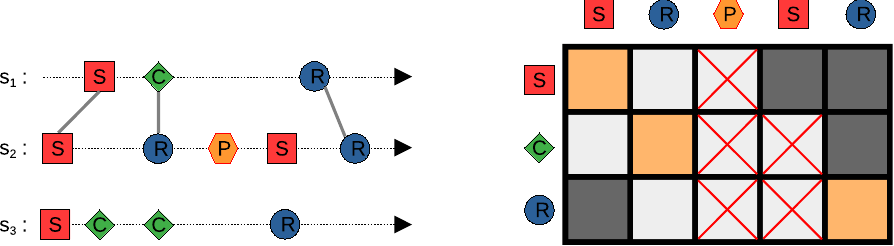}
\caption{Illustration of the timed sequences of Example~\ref{ex:ex1} (on the left) and the cost matrix for the alignment between $s_1$ and $s_2$ according to the drop-DTW (on the right). The alignment is illustrated by the grey bars on the left and by the colored cells in the matrix (on the right). The red-crosses illustrate that $P$ and $S$ has been dropped in this alignment, the cells at the corner are dark because they to no satisfy the constraint $\tau=3.5$. Clinical pathway interpretation: $S$ stands for surgery, $C$ consultation, $P$ physiotherapist session and $R$ radiotherapy session.}
\label{fig:il1}
\end{figure}

Example~\ref{ex:ex1} illustrates the representation of a timed sequence as a probabilistic timed sequence. Conversely, most of the probabilistic timed sequences can not be represented by a time sequence.

\begin{example}[Probabilistic representation of a timed sequence]\label{ex:ex1}
Let $\Sigma=\{S,R,C,P\}$ be an ordered set of event types. In the context of clinical care pathways, $S$ may represent a surgery event, $C$ a consultation, $P$ a physiotherapist session, and $R$ a radiotherapy session.
We define three timed sequences: $s_1=\langle (S,1), (C,2), (R,4.5)\rangle$, $s_2=\langle (C,0), (C,2), (P,3), (S,4), (R,5)\rangle$, and $s_3=\langle (S,0), (C,1), (C,2), (R,4)\rangle$ be three synthetic timed sequences illustrated in Figure~\ref{fig:il1}.
These sequences can be represented as probabilistic sequences:
$$\begin{array}{l}
s_1 = \langle([\mathsf{1,0,0,0}],1), ([\mathsf{0,0,1,0}],2), ([\mathsf{0,1,0,0}],4.5)\rangle\\
s_2 = \langle([\mathsf{0,0,1,0}],0), ([\mathsf{0,0,1,0}],2), ([\mathsf{0,0,0,1}],3), ([\mathsf{1,0,0,0}],4), ([\mathsf{0,1,0,0}],5)\rangle\\
s_3 = \langle([\mathsf{1,0,0,0}],0), [\mathsf{0,0,1,0}],1), (B,2), ([\mathsf{0,0,1,0}],2),([\mathsf{0,1,0,0}],4)\rangle
\end{array}$$
\end{example}

\section{Comparing Timed Sequences}\label{sec:metric}
This section introduces the drop-DTW for timed sequences. We first remind the definition of DTW and drop-DTW and then we adapt the latter for timed sequences.

\subsection{DTW and drop-DTW} \label{sec:dtw}

Dynamic Time Warping (DTW)~\citep{sakoe1978dynamic} computes the optimal alignment between two series subject to certain constraints. Here, we adapt these definitions to the case of timed sequences.
Let \(X = \langle(x_1,t^x_1), \dots, (x_N,t^x_N)\rangle\) and \(Z = \langle(z_1,t^z_1) \dots, (z_K,t^z_K)\rangle\) be two input sequences, where \(N, K\) are their respective lengths.
An alignment between two sequences is defined as a binary matrix, \(\pi \in \{0,1\}^{K \times N}\) where \(\pi_{i, j}=1\) if \(z_i\) is matched to \(x_j\), and \(0\) otherwise.
Matching an element \(z_i\) to an element \(x_j\) incurs a cost \(C_{i, j}\) between the \(i\)-th and \(j\)-th elements of the sequences.
The cost values are user-defined.
DTW finds the alignment \(\pi^*\) between sequences \(Z\) and \(X\) that minimizes the global cost:
\begin{equation*}\label{eq:dtw_objective}
\pi^* = \argmin_{\pi \in \mathcal{P}}\, \langle \pi, C\rangle = \argmin_{\pi \in \mathcal{P}} \sum_{i\in[K],\,j\in[N]} \pi_{i,j} C_{i,j},
\end{equation*}
where \(\langle \pi, C\rangle\) is the Frobenius inner product and \(\mathcal{P}\) is the set of all feasible alignments satisfying the following constraints: monotonicity, continuity, and end-point correspondence (\(\pi_{1,1}=\pi_{K,N}=1, \forall \pi \in \mathcal{P}\)).

drop-DTW~\citep{dvornik2021drop} extends the set of feasible alignments, \(\mathcal{P}\), to those adhering only to the monotonicity constraint. Consequently, unlike DTW, events can be dropped from the alignment process.
drop-DTW finds the optimal temporal alignment while allowing for the removal of certain elements.
To account for unmatched elements in the alignment cost calculation,~\citet{dvornik2021drop} introduce a new drop cost: \(\delta \in \mathbb{R}^+\).\footnote{In the original approach of Dvornik et al., the cost value depends on the event type. We use a constant value for the sake of simplification.}
The optimal matching can then be defined as follows:
\begin{equation*}\label{eq:dropdtw_objective}
\pi^* = \argmin_{\pi \in \overline{\mathcal{P}}}\, \langle \pi, C\rangle + \delta.\left(P_r(\pi) + P_c(\pi) \right)
\end{equation*}
where \(\overline{\mathcal{P}}\in\{0,1\}^K\) is the set of binary matrices satisfying only the monotonicity constraint, and \(P_c(\pi)\in\{0,1\}^N\) is a vector whose \(j\)-th element is \(1\) if \(\pi_{ :,j} = \bm{0}\) and \(0\) otherwise; \(P_r(\pi)\) is defined similarly, but along the rows.
The indices stand for \textit{r}ows and \textit{c}olumns.
Similar to DTW, drop-DTW can be efficiently evaluated using dynamic programming techniques.

\begin{example}[Drops representation in drop-DTW]
In Example~\ref{ex:ex1}, the alignment of the sequences $s_1$ and $s_2$ leads to skip the $P$ event and the second $S$ event (see Figure~\ref{fig:il1}). Then, the drops are represented as follows: $P_r=[1,1,1]$ and $P_c=[1,1,0,0,1]$.
\end{example}

\subsection{Adaptation of drop-DTW for timed sequences}\label{sec:adapt_dtw}
The proposed adaptation of drop-DTW focuses on proposing a metric between events and adding temporal constraints to the drop-DTW algorithm.
The Dynamic Time Warping algorithm requires defining a cost between two events to compare sequences. To take into account both time and events, we decided to assign weights to a time distance cost \(c_t\), and an event difference cost \(c_e\).
Having introduced the probabilistic representation of timed sequences, the distance between two events \(e_a=( \bm{d}_a=[d^1_a, \dots, d^n_a], t_a)\) and \(e_b=( \bm{d}_b=[d^1_b, \dots, d^n_b], t_b)\) is defined by:
\begin{equation*}d(e_a,e_b)=\sqrt{p_t.||\bm{d}_a-\bm{d}_b||_2^2 + p_e.(t_b-t_a)^2}
\end{equation*}
where \(p_t,p_e\in\mathbb{R}_+\) are weights of two terms: a distance between probabilistic events and a time gap. 
This metric implies that time events are in an Euclidean space of dimension \(n+1\). 
This is of particular interest in the next section, to define average events.

We can now adapt the metric between two timed sequences.
We have chosen to use drop-DTW for its ability to handle aberrant or missing events. 
Algorithm~\ref{algo:dropDTW} presents the method for calculating the proposed distance between two timed sequences. It takes as input the cost matrix $C$ that computes the cost of aligning two events using the distance \(d\). 
The algorithm finds the best alignment between sequence events to minimize the overall metric. The use of dynamic programming technique ensures the efficiency of the method. 
For the sake of computing efficiency, we also add constraints on the possible event pairings by allowing matches only between events that are close in time, \ie with a delay between their occurrence lower than a predefined threshold \(\tau\).

\SetKwInOut{Input}{Data}
\begin{algorithm}[t]
\caption{drop-DTW with proximity constraints}\label{algo:dropDTW}
\Input{$C\in \mathbb{R}^{M\times N}$: cost matrix; $\delta\in \mathbb{R^+}$: drop cost; $\sigma \in \mathbb{N^+}$, $\tau\in\mathbb{R^+}$: sequential/temporal proximity constraints}
\BlankLine
$D^+_{0,0} \gets 0$ ; $D^+_{i,0} \gets +\infty$ ; $D^+_{0,j} \gets +\infty${, $\forall i\in[M],\,j\in[N]$} \tcp*{Match table}
$D^-_{0,0} \gets 0$ ; $D^-_{i,0} \gets i \times \delta$ ; $D^-_{0,j} \gets j\times \delta${, $\forall i\in[M],\,j\in[N]$} \tcp*{Drop table}
{$D_{0,0} \gets 0$; $D_{i,0} \gets D^-_{i,0}$; $D_{0,j} \gets D^-_{0,j}$, $\forall i\in[M],\,j\in[N]$\tcp*{Optimal solution table} }
\For{$i=1$ to $N+1$}{
\For(\tcp*[f]{Sakoe-Chiba constraints}){$j=max(1,i-\sigma+1)$ to $min(m+1,i+\sigma)$}{
\If(\tcp*[f]{Temporal constraint}){$distdate(i-1,j-1) \leq \tau$}{
$D^+_{i,j} \gets C_{i-1,j-1} + min(D_{i-1,j},D_{i,j-1},D^+_{i-1,j-1})$ \tcp*{Alignment cost}
$D^-_{i,j} \gets \delta + D_{i,j-1}$ \tcp*{Drop cost}
$D_{i,j} \gets min(D^+_{i,j},D^-_{i,j})$ \tcp*{Final cost}
}\Else{
$D_{i,j} \gets +\infty$ \tcp*{Unsatisfied temporal constraints $\Rightarrow$ impossible match}
}
}
}
\Return{$D_{M,N}$, $D$}\;
\end{algorithm}

\begin{example}[Illustration of drop-cost usefulness]
Considering the sequences from Example~\ref{ex:ex1}, sequences $s_1$ and $s_3$ are very similar, but sequences $s_1$ and $s_2$ does not seem so different if we ignore some of the events. 
In the proposed alignment between $s_1$ and $s_2$, the drop-cost was lower than the cost of pairing $P$ in the sequence $s_2$ with another event of $s_1$. Thus, it was preferable to ignore this event. 
This is directly related to the definition of the distance between a physiologist session and other types of event. 
In the case of the second surgery event in $s_2$, it was too far in time from the one in $s_1$ according to the $\tau$ constraint. Then, against, it was preferable to ignore it in the final alignment.
\end{example}

\section{Construction of an Average Timed Sequence}\label{sec:average}
Computing an average timed sequence consistent with the notion of similarity is essential for clustering algorithms. 
In the following, we propose a method inspired by the DBA algorithm~\citep{petitjean_summarizing_2012} to create average sequences in terms of drop-DTW, taking into account the possibility of removal.
DBA addresses the problem of generating an average time series in terms of DTW -- which does not hold distance properties. 
In this case, the generalized average is defined as a time series that minimizes the global inertia (see Equation~\ref{eq:inertia}).

DBA iteratively refines a (randomly initialized) series to minimize its quadratic distance (DTW) with respect to the series to be averaged. At each iteration, the algorithm performs ``vertical'' centroids of the aligned elements of the time series. The definition of a ``vertical'' centroid corresponds to averaging series real values. 
Our approach enables the adaptation to timed sequences thanks to the representation of events as distributions of types.

\subsection{Average Timed Sequences}
Let $\mathcal{S}=\{s_1,\dots,s_m\}$ be a set of timed sequences, Algorithm~\ref{algo:TSR} details the steps for generating an average timed sequence inspired by the principle of \textit{vertical} averages of DBA~\citep{petitjean_global_2011}. 
In this algorithm, $\mathcal{C}^{(k)}$ denotes the current average sequence at the $k$-th iteration.
At line 1, the current average time sequence, $\mathcal{C}^{(0)}$ is initialized with a timed sequence chosen from $\mathcal{S}$. Then, this current average is processed into the outer loop which repeats $maxit$ times the alignment and update steps. 
The alignment step (lines 6-11) calculates the alignment by drop-DTW between $\mathcal{C}^{(0)}$ and each of the other sequences. The alignment to one sequence is obtained line 8 from the distance matrix computed the drop-DTW. We refer the reader to the original DTW algorithm to get the details on how to compute an alignment from a distance matrix.  
Lines~9 to~11 collect, on the one hand, the list of paired events in the sequences of $\mathcal{S}$ per event id, $e$, in $\mathcal{E}_e$ and, on the other hand, their dates per event id, $e$, in $\mathcal{T}_e$. 
These lists are then averaged lines~12-13. This is the update step. This vertical average of events is constructed as the histogram of event distributions. 
Lastly, line 14 creates a new average sequence which can be refined in a subsequent iteration until convergence. In our implementation, we assume convergence after a fixed number of iterations but a convergence criteria could be implemented to prevent useless iterations.

Similarly to DBA, events from probabilistic sequences are represented in a vector space. Thus, the new average sequence is always closer (according to drop-DTW) to the sequences it averages, and the algorithm converges well (see Section~\ref{sec:proof}). Additionally, this algorithm preserves the drop-DTW's ability to allow temporal shifts and handle missing or aberrant events.

The choice of the timed sequence in line 1 is crucial for the result. 
We choose to randomly select a sequence of the longest length. 
We made this choice because our algorithm decreases the length of the current average timed sequence, but it can not increase it. Note that there are some smarter initialization strategies for k-means like algorithms that could improve the convergence, for instance K-means++~\citep{arthur2006k}, or recent provably good seeding methods \citep{bachem2016fast}.

\begin{algorithm}[t]
\caption{TSR: Generation of an Average Timed Sequence}
\label{algo:TSR}
\Input{$\mathcal{S} = \{s_1,s_2,...,s_m\}$: list of timed sequences; $maxit\in\mathbb{N}$: maximum number of iterations}
$\mathcal{C}^{(0)} \gets alea(\mathcal{S})$\;
$k\gets 0$\;
\While{k<maxit}{
$\mathcal{E} \gets [\emptyset,\dots,\emptyset]$ \tcp*{sets of length $|r|$}
$\mathcal{T} \gets [\emptyset,\dots,\emptyset]$ \tcp*{idem}
\For{$i = 1,\dots,m $} {
$\mu, D \gets DropDTW(\mathcal{C}^{(k)},s_i)$ \tcp*{calculate the distances matrix of drop-DTW between $\mathcal{C}^{(k)}$ and $s_i$}
$\pi^* \gets GetAlignement(D)$ \tcp*{calculate the best alignment from $D$}
\For{$(i_r,i_s) \in \pi^*$}{
$\mathcal{E}_{i_r} \gets \mathcal{E}_{i_r} \cup  \{s_i[i_s].\bm{d}\}$ \tcp*{add event from $s_i$ aligned at $i_r$}
$\mathcal{T}_{i_r} \gets \mathcal{T}_{i_r} \cup  \{s_i[i_s].t\}$ \tcp*{add date from $s_i$ aligned at $i_r$}
}
}
$\tau \gets \left\{\tau_{j} \,\mid\, \forall j\in[|\mathcal{C}^{(k-1)}|],\; \tau_{j} = \frac{1}{|\mathcal{T}_{j}|}\sum_{t\in\mathcal{T}_{j}}t ,\, |\mathcal{T}_{j}|\neq 0\right\}$ \tcp*{calculate vertical averages}
$ \mathcal{H} \gets \left\{\bm{h}_j \,\mid\, \forall j\in[|\mathcal{C}^{(k-1)}|],\; \bm{h}_{j} = \frac{1}{|\mathcal{E}_{j}|}\sum_{\bm{d}\in\mathcal{E}_{j}}\bm{d},\, |\mathcal{E}_{j}|\neq 0 \right\}$ \;
$\mathcal{C}^{(k+1)} \gets (\mathcal{H}, \tau)$\tcp*{refined average timed sequence}
$k\gets k+1$\;
}
\Return{$\mathcal{C}^{(k)}$}\;
\end{algorithm}

\begin{figure}[tb]
\centering
\includegraphics[width=\textwidth]{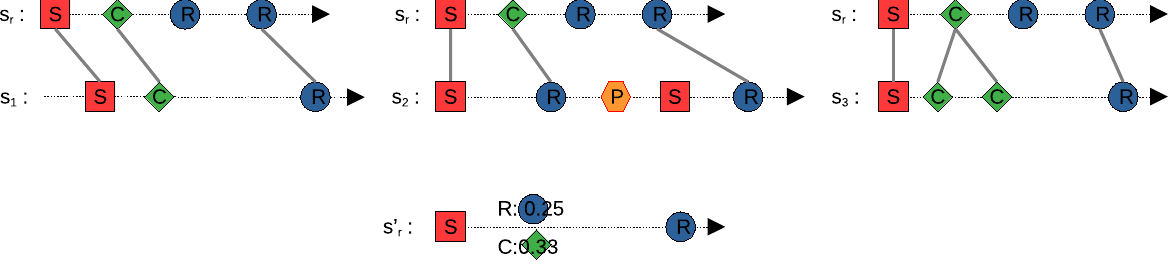}
\caption{Illustration of one averaging step. At the top, the alignment of the current average $s_r$ is computed with each sequence. Then, the ``vertical'' average of the set of events paired to one $s_r$ event yields an average probabilistic event in $s'_r$ (at the bottom).}
\label{fig:averaging}
\end{figure}

\begin{example}[Iteration of the averaging algorithm]
Figure~\ref{fig:averaging} illustrates one iteration of average calculation for the three sequences in Example~\ref{ex:ex1}. 
It assumes that we already have a current average sequence $s_r$ (a surgery, then a consultation and two radiotherapy sessions). 
The figure depicts the alignments by drop-DTW of $s_r$ with each sequence of Example \ref{ex:ex1}. 
The gray lines represents the alignments (line 8). 
$s'_r$ is the updated average sequence obtained by the lines 11-12 of Algorithm~\ref{algo:TSR}. 
As the $C$ event has been paired with both $C$ and $R$ events, the second probabilistic event of $s'_r$ combines the two types of events (proportionally). In addition, the first $R$ event of $s_r$ is removed from $s'_{r}$ because it is paired with no event. The other events are purely paired with similar event types, then the probabilistic event is a pure $S$ or $R$ in $s'_r$ but their time positions have been adjusted with respect to the positions of the new aligned events. 
\end{example}

\subsection{Convergence of the Algorithm}\label{sec:proof}
In this section, we prove the theoretical convergence of the proposed algorithm.

The algorithm~\ref{algo:TSR} iteratively refines an average sequence for a set of sequences, $\left\{s_i\right\}_{i\in[m]}$. 
At step $k$, the inertia of the average sequence is defined as:
\begin{equation}\label{eq:inertia}
I^{(k)} = \frac{1}{m}\sum_{i\in[m]} DropDTW\left(\mathcal{C}^{(k)},s_i\right)
\end{equation}
The inertia represents the average error incurred by using $\mathcal{C}^{(k)}$ as the representative of the set of sequences $\left\{s_i\right\}_{i\in[m]}$.
To demonstrate the convergence, we prove that $I^{(k+1)}$ is a decreasing suite, i.e., $I^{(k+1)}\leq I^{(k)}$, for all $k\in\mathbb{N}$.

The $i$th term of the inertia is
\begin{equation*}
DropDTW\left(\mathcal{C}^{(k)},s_i\right)=\min_{\pi} \sum_{(x,y)\in\pi}d\left(\mathcal{C}^{(k)}[x],s_i[y]\right) + \delta.\left(P_r(\pi) + P_c(\pi) \right)
\end{equation*}
where $d$ is the metric between timed events (see Section~\ref{sec:adapt_dtw}) and $\pi$ plays the same role as in section~\ref{sec:dtw}. It represents the alignment between two sequences. 
In the remainder, we denote by $\pi^{(k)}_i$ the optimal alignment of sequence $s_i$ with sequence $\mathcal{C}^{(k)}$, i.e., the alignment that minimizes the sum of costs. 

Let us now introduce $q_i^{(k+1)}=\sum_{(x,y)\in\pi^{k}_i}d\left(\mathcal{C}^{(k+1)}[x],s_i[y]\right) + \delta.\left(P_r(\pi^{k}_i) + P_c(\pi^{k}_i) \right)$, for all $i\in[m]$ and $k\in\mathbb{N}^*$.
Providing $\pi^{k+1}_i$ is the optimal alignment, we have:
\begin{equation*}
\sum_{(x,y)\in\pi^{k+1}_i}d\left(\mathcal{C}^{(k+1)}[x],s_i[y]\right)+ \delta.\left(P_r(\pi^{k+1}_i) + P_c(\pi^{k+1}_i) \right) \leq q_i^{(k+1)} 
\end{equation*}

It is worth noticing that we retrieve the inertia of the $(k+1)$-th average sequence $I^{(k+1)}$ by summing the left-hand terms over $i$, and then: 
\begin{equation}\label{eq:res1}
I^{(k+1)}\leq \frac{1}{m}\sum_{j\in[m]} q_i^{(k+1)} 
\end{equation}

Let us not have a look at the sum over $i$ of the $q_i^{(k+1)}$. 
This sum can be rewritten as follows:
\begin{eqnarray*}
\frac{1}{m}\sum_{j\in[m]} q_j^{(k+1)} & = & \frac{1}{m}\sum_{j\in[m]}\sum_{(x,y)\in\pi^{k+1}_j}d\left(\mathcal{C}^{(k)}[x],s_j[y]\right)+ \delta.\left(P_r(\pi^{k+1}_j) + P_c(\pi^{k+1}_j) \right) \\
& = &\sum_{x}\frac{1}{m}\sum_{j\in[m]}\sum_{\left\{y\mid(x,y)\in\pi^{k+1}_j\right\}}d\left(\mathcal{C}^{(k)}[x],s_j[y]\right)+ \delta.\left(P_r(\pi^{k+1}_j) + P_c(\pi^{k+1}_j) \right)
\end{eqnarray*}

Let us now have a look at the construction of $\mathcal{C}^{(k+1)}[x]$ in the algorithm (see lines 11-13). It is the barycenter of all elements $s_j[y]$ that have been associated with $\mathcal{C}^{(k)}[x]$ by the alignment $\pi^{k+1}_j$. 
Thanks to the property of barycenters in an Euclidean space, we have for any $\bm{v}$:
\begin{equation}\label{eq:res_inter}
\frac{1}{m}\sum_{j\in[m]}\sum_{\left\{y\mid(x,y)\in\pi^{k}_j\right\}}d(\bm{v},s_j[y]) \geq \frac{1}{m}\sum_{j\in[m]}\sum_{\left\{y\mid(x,y)\in\pi^{k}_j\right\}}d(\mathcal{C}^{(k+1)}[x],s_j[y])
\end{equation}

Considering that:
\begin{equation*}
I^{(k)} = \sum_{x}\frac{1}{m}\sum_{j\in[m]}\sum_{\left\{y\mid(x,y)\in\pi^{k}_j\right\}}d\left(\mathcal{C}^{(k)}[x],s_j[y]\right)+ \delta.\left(P_r(\pi^{k}_j) + P_c(\pi^{k}_j) \right) 
\end{equation*}
and applying inequation~\ref{eq:res_inter} with $\bm{v}=\mathcal{C}^{(k)}[x]$, it ensues:
\begin{eqnarray}\label{eq:res2}
I^{(k)} &\geq &\sum_{x}\frac{1}{m}
\sum_{j\in[m]}\sum_{\left\{y\mid(x,y)\in\pi^{k}_j\right\}}d(\mathcal{C}^{(k+1)}[x],s_j[y])+ \delta.\left(P_r(\pi^{k}_j) + P_c(\pi^{k}_j) \right) \nonumber\\
&\geq & \frac{1}{m}\sum_{j\in[m]} q_j^{(k+1)}
\end{eqnarray}

In the case $\left\{y\mid(x,y)\in\pi^{k}_j\right\}=\emptyset$, this means that for all $s_i$, there is an empty row at line $x$ of the alignment with $\mathcal{C}^{(k+1)}$. For instance, the first $R$ event in Figure~\ref{fig:averaging} is assigned to none event. Then, this event contributes to the inertia only via a drop-cost.
Removing such an event from $\mathcal{C}^{(k+1)}$  turns out to remove an empty row for all sequences $s_j$, and thus subtracts one to all $Pr(\pi^{k+1}_j)$. Then, it also decreases the true values of $q_j^{(k+1)}$.

Finally, with equations~\ref{eq:res1} and~\ref{eq:res2}, we conclude that $I^{(k+1)}\leq I^{(k)}$. This proves that Algorithm~\ref{algo:TSR} actually refines the average timed sequence and converges. Nevertheless, it does not guarantee to find the global minimum of inertia. It only ensures to find a local minimum.

\subsection{Parameters}
The proposed method for averaging time sequences has two types of parameters: the parameters of the drop-DTW metric ($\tau\in\mathbb{R}$, $p_e\in[0,1]$, $p_t\in[0,1]$, $\delta\in\mathbb{R}$) and the parameters of the Algorithm~\ref{algo:TSR} (maximum iterations for drop-DTW and iterative averaging). 

The proposed metric is mainly parameterized by the ratio $\frac{p_t}{p_e}$ and the drop-cost $\delta$.
We explicitly explain here how to determine these values relative to the unique threshold value $\tau$. Thus, the expert only has one parameter to set up: $\tau$, which can be seen as a threshold of indifference delay. Above this delay, two events cannot be considered as similar, regardless of their types. An expert can set up $\tau$ based on this intuition.

The ratio $\frac{p_t}{p_e}$ can be inferred from the delay $\tau$ by the following reasoning. 
For events located within the time interval $[t - \tau,t+ \tau]$, the importance of the event type should be greater than the importance of time. We must then choose $p_e$ to be at least $\tau^2$ times greater than $p_t$.
Without lost of generality, the value of $p_t$ can be set at $1$ by default. 
The drop-cost is also derived from $\tau$, i.e., the delay from which there is no longer any interest in associating two events. Thus, we propose to define the drop-cost as:
$$\delta = \frac{1}{\tau^2}(\tau+ t_{max}) +1$$
where $t_{max}$ is the maximum time distance to consider.

Note that this suggested values are only a guideline to initiate a set up for the metric. Nonetheless, it has to be adjusted and tested on real data.

\section{Clustering of Timed Sequences}\label{sec:clustering}

The proposed metric between timed sequences and the averaging technique are used to adapt two classical clustering techniques: a hierarchical ascendant clustering (HAC) and a K-means algorithm~\citep{hartigan1975clustering}. These two algorithms require to predefine the expected number of clusters, $K$.\footnote{Model selection based on the inertia of the clustering may be applied to select the optimal number of clusters.}

\HierASTiSeq denotes the HAC algorithm for timed sequences. The principle of the algorithm is to iteratively aggregate the two closest timed sequences according to the drop-DTW. 
A group of two timed sequences is then averaged using Algorithm~\ref{algo:TSR}, and the average timed sequence replaces the two former ones. 
The algorithm terminates when it reaches the expected number of clusters.

The $K$-means algorithm also proceeds to refine iteratively $K$~clusters. K-means algorithm has two main steps: first the sequences of each cluster are averaged (Algorithm~\ref{algo:TSR}). The computed averages are the representatives of their clusters. Then, new clusters are made by assigning optimally each sequence to one of the $K$~representatives. 

\section{Experiments and Results on Synthetic Data}\label{sec:results}
In this section, we present experiments on synthetic data. 
We conducted several tests to evaluate the effectiveness of the method under various scenarios and to study the effect of the parameters. 

\subsection[Effects of the time/event Ratio]{Effects of the $\frac{p_t}{p_e}$ Ratio}

For this first experiment, we created two model timed sequences that differ in one or more of the dimensions we want to study: sequence length, event nature, and event dates. Therefore, there are $8=2^3$ possible variations.
A dataset consists of 15 sequences for each of the 8 possible variations and the base sequence (thus $135=9\times 15$ sequences in total). The dates of these timed sequences are obtained and made random by adding uniform noise to the date.
The objective is to evaluate the accuracy of identifying the 9 categories of sequences.

We tested two sets of parameters: first, we chose the ratio $\frac{p_t}{p_e} = \frac{1}{9}$, with $\delta=+\infty$. With this configuration, the cost of an event is more important than the time cost for 3 days around the event.
The second set of parameters used was~$\frac{p_t}{p_e} = \frac{1}{400}$, still with~$\delta=+\infty$. This time, the difference cost between events is more important than the time cost over the first 20 days.

In the case of the first configuration, we observe that all clusters generated by \HierASTiSeq were correctly identified (Kappa score of $1$). However, in the case of the second configuration, we observe a degradation of results with a Kappa score of $0.62$. This degradation was expected because the maximum time distance observed between events is around 20. Consequently, the notion of time is practically never taken into account by the metric.

This experiment thus shows that the choice of parameters is crucial but can be made by taking into account the distance between events of the sequences.

\subsection{Effect of Drop-Cost}
We can then test the effect of the drop-cost on a synthetic dataset, with and without drop-cost. We consider 3 models of sequences. We generate 15 concrete sequences from each of them by randomizing the date of the events\footnote{We added a centered normal random noise with a standard deviation of 1.}. The sequences of the first and the third types are similar, except that the sequences of the third type have an aberrant event.
\begin{enumerate}
\item $\langle(D, 0.0) (E, 3.0) (F, 5.0) (D, 7.0) (D, 10.0) (E, 12.0) (F, 15.0)\rangle$
\item $\langle(E, 2.0) (A, 4.0) (D, 8.0) (C, 12.0)\rangle$
\item $\langle(D, 0.0) (E, 3.0) (F, 5.0) (D, 7.0) (D, 10.0) (E, 12.0) (F, 15.0) (D, 1500.0)\rangle$
\end{enumerate}

\begin{table}[t]
\centering
\begin{tabular}{ccc}

& \multicolumn{2}{c}{\textit{\small Predicted}} \\
\hline
\multirow{3}{*}{\rotatebox[origin=l]{90}{\small Real}} & 0 & 15\\
& 15 & 15\\
\hline
\end{tabular}
\hspace{2cm}
\begin{tabular}{ccc}
&\multicolumn{2}{c}{\textit{\small Predicted}} \\
\hline
\multirow{3}{*}{\rotatebox[origin=l]{90}{\small Real}} & 15 & 0\\
& 0 & 30\\
\hline
\end{tabular}
\caption{Confusion matrices with supplementary events. On the left: $p_t = \frac{1}{9}$, $p_e = 1$,$\delta = +\infty$; On the right: $p_t = \frac{1}{9}$, $p_e = 1$, $\delta = 4$}
    \label{tab:conf_matrices_drop_add}
\end{table}

Intuitively, we want to cluster together all sequences of types 1 and 3.
We apply \HierASTiSeq without drop-cost and obtain the confusion matrix in Table~\ref{tab:conf_matrices_drop_add} on the left, with a Kappa score of $-0.5$ (disagreement). We observe that the algorithm associated instances of sequence 1 with instances of sequence 2.

Adding drop-cost coinciding with the distances between the sequences allows us to obtain the confusion matrix in Table~\ref{tab:conf_matrices_drop_add} on the right, with a Kappa score of~$1$ (total agreement). The drop cost allows neglecting the last event considered aberrant, and the results match expectations.

\begin{table}[t]
\centering
\begin{tabular}{ccc}

& \multicolumn{2}{c}{\textit{\small Predicted}} \\
\hline
\multirow{3}{*}{\rotatebox[origin=l]{90}{\small Real}} & 0 & 15\\
& 15 & 15\\
\hline
\end{tabular}
\hspace{2cm}
\begin{tabular}{ccc}
&\multicolumn{2}{c}{\textit{\small Predicted}} \\
\hline
\multirow{3}{*}{\rotatebox[origin=l]{90}{\small Real}} & 15 & 0\\
& 0 & 30\\
\hline
\end{tabular}
\caption{Confusion matrices with missing events. On the left: $p_t = \frac{1}{9}$, $p_e = 1$,$\delta = +\infty$; On the right: $p_t = \frac{1}{9}$, $p_e = 1$, $\delta = 4$}
    \label{tab:conf_matrices_drop_remove}
\end{table}

We also conducted a similar experiment with missing values. This time, the model sequences are as follows:
\begin{enumerate}
\item $\langle(D, 0.0),(E, 3.0),(F, 5.0),(D,7.0)\rangle$
\item $\langle(E, 2.0),(A, 4.0),(E, 7.0),(D, 8.0),(D,10.0),(E,12.0)\rangle$
\item $\langle(D, 0.0),(E, 2.0),(E, 3.0),(F, 5.0),(D,7.0),(F,9.0),(A,13.0)\rangle$
\end{enumerate}

We then notice that all events dated from sequence 1 are present in sequence 3, but they are all different from sequence 2. We would therefore like to cluster sequences from 1 and 3 together. Keeping the same parameters as before. Without drop-cost, we obtain the confusion matrix in Table~\ref{tab:conf_matrices_drop_remove} on the left. This result corresponds to expectations because the sequences from 2 and 3 are then associated as they are closer in time.
However, with $\delta=4$, we obtain the confusion matrix in Table~\ref{tab:conf_matrices_drop_remove}. The table on the right, as expected, shows that the events furthest in time are no longer associated.

The tests on synthetic data lead to the expected results, thus validating the effectiveness of clustering in the situations described. We now continue studying the method with an experiment on real data to evaluate its ability to identify relevant clusters.

\section{Application to Real-World Care Pathways}\label{sec:usecase}
In this section, we apply the proposed method to real-world care pathways and compare our results with those of Sequence State Analysis (SSA)~\citep{gabadinho_analyzing_2011}. SSA, particularly the TraMineR tool, is commonly used for categorizing care pathways~\citep{roux2019use,le2019categorical}.

This experiment is conducted on the data warehouse of the Greater Paris Hospital (AP-HP). 
We identified a population of $3\,311$ patients having a stay with a lung resection surgery. 
We collected events that occurred within 90~days around a lung resection surgery as well as the date of death if applicable. 
Events further away are clinically less interesting to study.
For clinical reasons, we also decided to keep only the first occurrence of chemotherapy for each patient. We made the same choice for radiotherapy and immunotherapy. 

The parameters chosen are \( K=5 \) clusters, \( \tau = 7 \), \( t_{\text{max}} = 16 \), and \( \delta = 1 \). 
For choosing the number of clusters, different values were tested. By increasing the number of clusters, it is mainly small-sized clusters that split. Therefore, there may not necessarily be an interest in increasing this number.
\( \tau \) is set to 7 days because, in our dataset, it may be interesting to associate the same types of events that occurred within a week of each other. Finally, based on the dataset, we consider that there is no interested in pairing two events more than 16 days apart.

For this experiment, we used the servers from the Paris University Hospital made of AMD EPYC 7713 processors. Our algorithms are neither parallelized nor distributed. However, the implementation of softDTW by \cite{tavenard2020tslearn} can utilize multiple cores. During our experiments, we requested only one core from the servers.

\begin{figure}[t!]
    \centering
    \includegraphics[width=.32\textwidth]{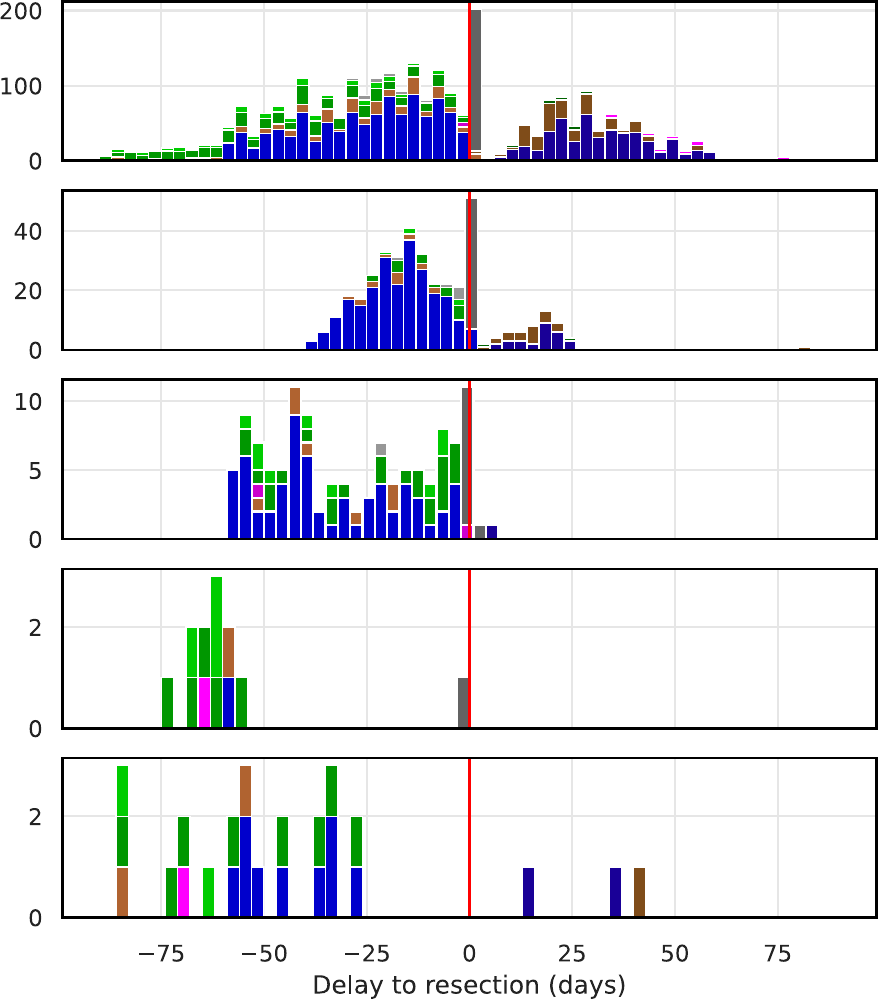}
    \includegraphics[width=.32\textwidth]{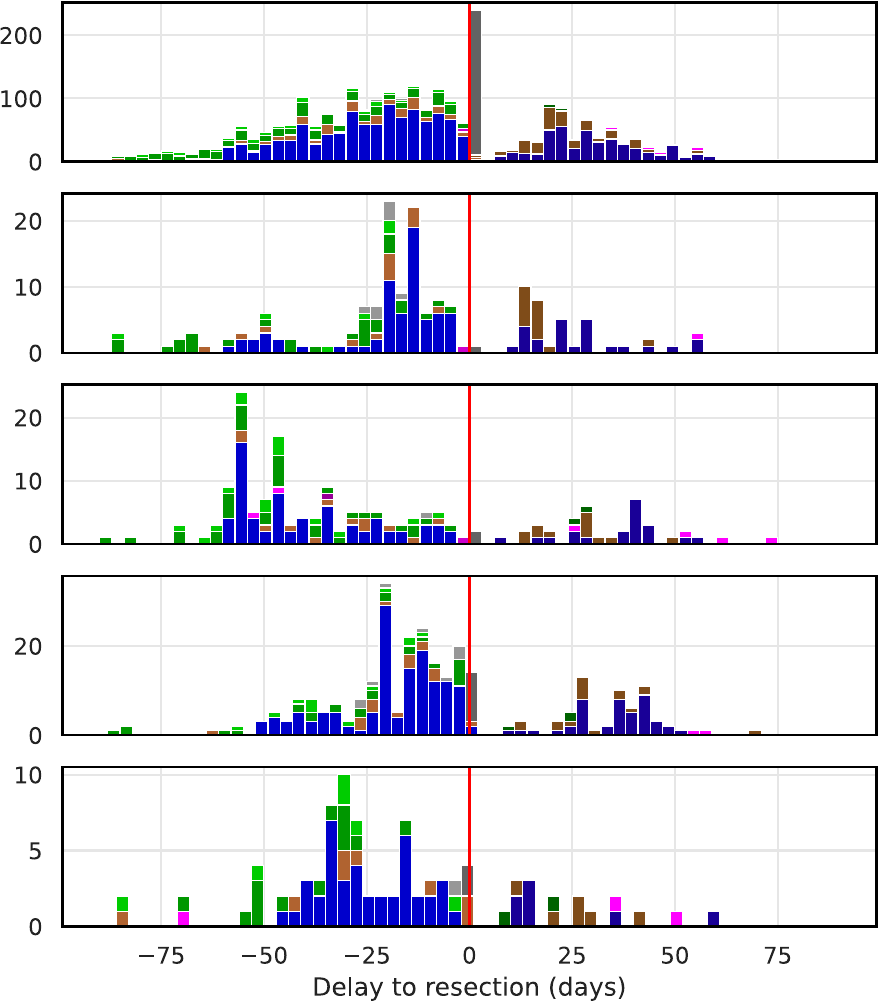}
    \includegraphics[width=.32\textwidth]{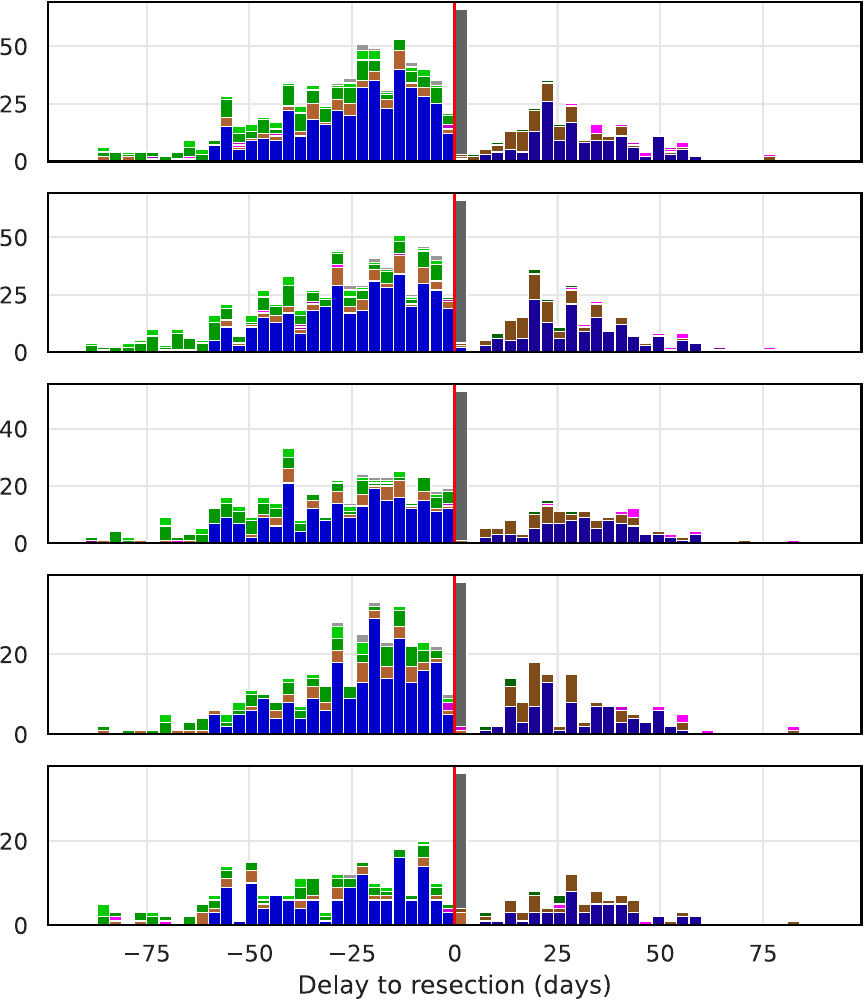}
    \caption{Clusters obtained by the clustering methods, from left to right: \HierASTiSeq, K-means and TraMineR. Clusters are ordered row-wise by their size. 
    Each cluster is represented by the histogram of event types over time. 
    Each color corresponds to a specific type of event (see Figure~\ref{fig:5cluster_K-means_clustiseq} for color legend). The higher the bar, the more events there are at the given time across all patients. The vertical red line indicates the index-date of the resection (0 delay). }
    \label{fig:5cluster_500patients_clustiseq}
\end{figure}

\subsection{Comparison with TraMineR}
In this section, we selected randomly 500~patients among the cohort of patients to compare the three clustering approaches: \HierASTiSeq, K-means with drop-DTW and TraMineR. Regarding the computing time of \HierASTiSeq and TraMineR, this comparison is conducted only on a small subset of patients.
TraMineR requires converting timed sequences into time series.
As TraMineR is sensitive to sparse series, we increased the time sampling from~1 to 3~days for calculations, and we added the empty event corresponding to the absence of data over a period of 3~days. TraMineR uses the hierarchical clustering (ward criteria) and on optimal matching metric between state sequences.

Figure~\ref{fig:5cluster_500patients_clustiseq} corresponds to the $K=5$ clusters for the three methods. Each cluster is represented as an histogram of events over time. 
One can first notice that TraMineR provides clusters that look very similar to each others. It is difficult to identify specificities for each cluster. Contrary to the other two methods, based on our drop-DTW metric, the sizes of the clusters are more homogeneous. 
For \HierASTiSeq and K-means, there is one large cluster and several small clusters that gather patient specific timed events. The two large clusters (cluster~1 for \HierASTiSeq and cluster~1 for K-means) looks very similar in distribution of events. 
\HierASTiSeq identifies two clusters with specific behaviors: the cluster~3 has no post-operative events. This may correspond to patients that had a surgery in the hospital but their follow-up in another medical structure; the cluster~2 are patients with short delays between pre- or post-operative care events around the surgery. 
The clusters~3 and~4 identified by K-means also gather patients with a late pre-operative consultation (wrt the resection). According to the distributions, these patients have two pre-operative consultations: in cluster~3, the first pre-operative consultation occurs earlier than in cluster~4; and cluster~4 holds patients having a late post-operative consultation. 

We conclude from this experiment that the use of drop-DTW yields clusters of patients with more specific timed sequences than TraMineR. For this dataset, the latter method does not seem to be able to reveal interesting clusters of patients. We performed a quantitative (confusion matrices) and qualitative analysis. The analysis of confusion between \HierASTiSeq and TraMineR shows that there is no correspondence between the clusters. This suggests that the metrics are semantically different. 
In this case, the semantics of drop-DTW on timed sequence seems more meaningful, but it does not mean that it would be the case for all datasets. 

It is worth noting that results of K-means with drop-DTW may significantly differ across different runs due to the randomness of the cluster seeds. 
\HierASTiSeq and TraMineR are more deterministic, but the iterative computation of drop-DTW or optimal matching may introduce some randomness. 
Nonetheless, the hierarchical clustering strategy of \HierASTiSeq and TraMineR is resource demanding. This prevents applying them on the entire dataset. This was possible only with K-means and takes only a few minutes.

\begin{figure}[tb]
    \centering
    \includegraphics[trim={1.25cm 1.cm 1.25cm 2cm},clip, width=.75\textwidth]{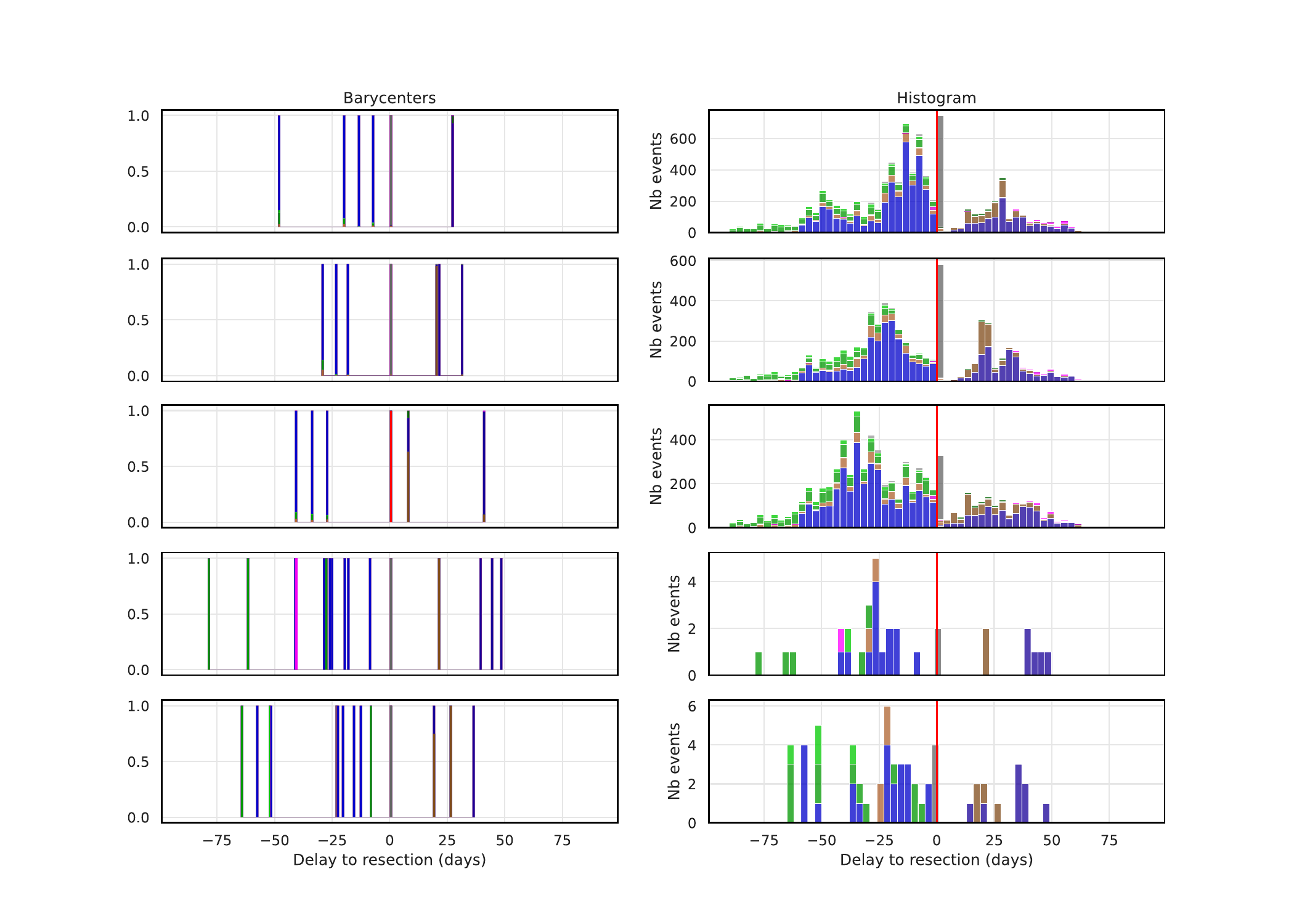}
    \includegraphics[width=0.2\textwidth]{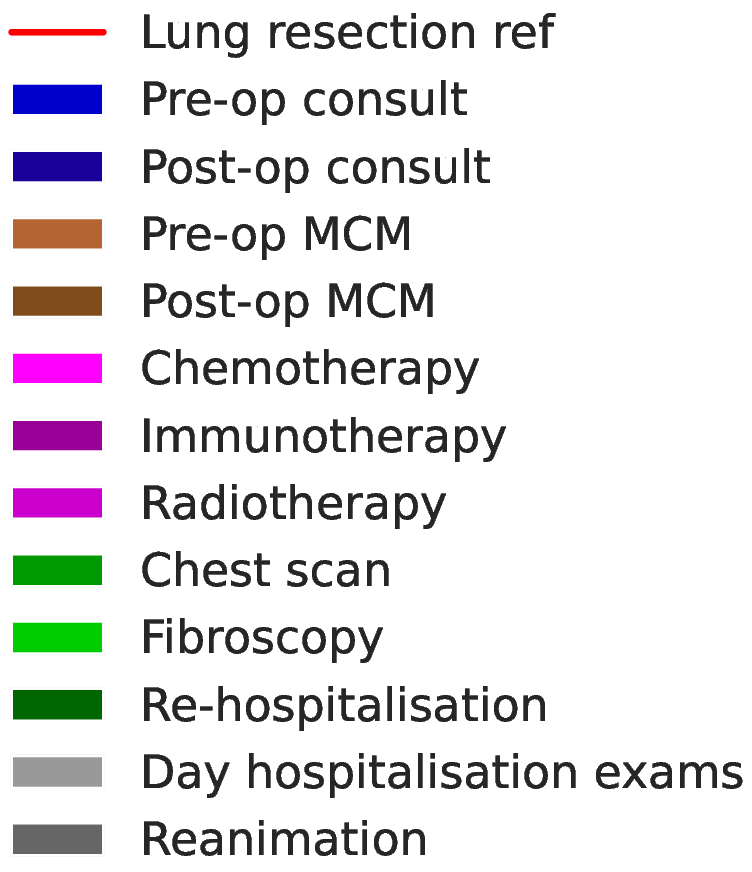}
    \caption{Clusters of care-pathways identified by K-means with drop-DTW on~{$3\,311$} patients. Each row is a cluster with the representation of the barycenter on the left and the representation of the events distribution over time on the right. Each color corresponds to a specific type of event.}
    \label{fig:5cluster_K-means_clustiseq}
\end{figure}

\subsection{Cluster Analysis of K-means with drop-DTW}
In this section, we present the results of the K-means clustering with drop-DTW metric applied on the entire cohort. Figure~\ref{fig:5cluster_K-means_clustiseq} illustrates the $K=5$ clusters. 
Each cluster is depicted with its barycenters and the distribution of events. 

Let us first comment on the clusters based on their event distributions. Three main clusters have been identified (clusters~1, 2 and 3 and share almost the same size (about 1000 patients). The two other clusters are very small (numbered~4 and~5). These small clusters of patients seem to be outliers. 
Similarly to the previous results, the K-means identified clusters of patients with different delays between the pre-operative consultations and resection. The cluster 1 corresponds to patients having a small delay between these events and possibly two pre-operative consultations. These patients possibly have a late stage of the disease and need a quick surgery. We made this hypothesis because their survival at 6~months is significantly lower than the two others clusters. 
The cluster~2 and cluster~3 differ mainly in the delay between the surgery and the post-operative consultations. The cluster~3 also holds patients without exams the day of the resection. 

The same observations can be made based on the analysis of the barycenters. 
We remind that a barycenter is a probabilistic timed sequence. Each event is represented by a distribution of event's types. On the figure, each bar is an event located at its time of occurrence. A bar stacks the probabilities per event's type. We remark that most of the events have one main type. The event at the 10th day for cluster~3 is a mixture of mainly two events types: post-operative consultation or post-operative multi-disciplinary consultation meeting (MCM). This possible confusion may be explained by the imprecision of the records. MCM are identified by consultation reports. If the report is not correctly tagged as an MCM, then it is simply a consultation. 
Then, the flexibility of the metric makes it somehow robust to possible labeling errors in the data.

We can also notice that there is a clear alignment between the barycenters' events and the event distribution. This illustrates that the averaging method, based on drop-DTW provides a meaningful average of a collection of complex timed sequences. 
More specifically, comparing the clusters through the decay between consultations and surgery leads to the same conclusions as before.

Lastly, one can note that there are a lot of repeated events in the barycenter. Furthermore, the presence of repeated events in the barycenter can be misleading, as it could improve either indicate a repetition of events or alternative delays for a single event due to possible drops.

\section{Conclusion}
We proposed a method for clustering timed sequences inspired by existing techniques for time series. Thus, the use of timed sequences allows co-occurring events, irregular time sampling, and sparse events. All these advantages are very interesting from a clinical point of view for describing care pathways.
We showed that by using drop-DTW, the clustering algorithm is less sensitive to missing and aberrant values when drop-DTW is used. The result is interesting for application to real-world care pathways. Then, we tested our method on real data from the OPTISOINS study for our approach compared to TraMineR. Based on the initial results, drop-DTW seems to be clinically promising for identifying average pathways.

However, the large number of parameters is a first weakness of the method. Regarding the number of parameters, our effort focused on a guide for choosing parameter values, which the experiments tend to validate. 
In perspective, a more in-depth study of the effect of parameters still needs to be carried out. Additionally, we have to validate the clinical interest of average pathways by clinicians.

Finally, the number of computations of a drop-DTW required during the clustering of timed sequences may be very high and may lead to long computation times for large datasets. Some strategies that could make use of distributed resources (grid computing) or fast approximate of the metric could significantly the practical usability of our method.

\paragraph{Acknowledgments}
The authors thanks gratefully Dr. Yen-Lân NGuyen from Department of Anaesthesia and Intensive Care Medicine, AP-HP, Cochin Academic Hospital, for her time and expertise she provided during this project. 
Part of the research presented in this article is funded by the AP-HP Foundation, as part of the AI-RACLES Chair, and has received approval from the Scientific and Ethical Committee of the AP-HP CDW (CSE-20-01-OPTISOINS).

\bibliographystyle{plain}
\bibliography{biblio}

\end{document}